\documentclass[a4paper, 10pt, conference]{ieeeconf}      
\usepackage{FG2024}
\usepackage{graphicx}
\usepackage{multirow}
\usepackage{array}
\usepackage{subcaption}
\usepackage{xurl}
\usepackage{amsmath}
\usepackage{amssymb}
\usepackage{booktabs,cite}

\newcommand{\secref}[1]{\S\ref{#1}}
\graphicspath{{./figures/}}

\FGfinalcopy 

\IEEEoverridecommandlockouts                              
\usepackage[pagebackref,breaklinks,colorlinks]{hyperref}

\newcommand{\modelname}{DiCTI}

\title{\LARGE \bf
\modelname: Diffusion-based Clothing Designer via Text-guided Input
}

\author{\parbox{16cm}{\centering
    {Ajda Lampe$^2$, Julija Stopar$^1$,  Deepak K. Jain$^3$, Shinichiro Omachi$^4$, Peter Peer$^2$, Vitomir Štruc$^1$}\\\vspace{2mm}
    {\normalsize
    $^1$ University of Ljubljana, Faculty of Electrical Engineering, Ljubljana, Slovenia\\
    $^2$ University of Ljubljana, Faculty of Computer and Information Science, Ljubljana, Slovenia\\
    $^3$ Dalian University of Technology, China\\
    $^4$ Tohoku University, Graduate School of Engineering, Sendai, Japan}}
    \thanks{This was supported by the Slovenian national research agency ARIS in research project J2-2501, and ARIS programmes P0-0250 and P2-0214.}
}

\usepackage{fancyhdr}
\begin{document}

\ifFGfinal
\thispagestyle{empty}
\pagestyle{empty}
\else
\author{Anonymous FG2024 submission\\ Paper ID \FGPaperID \\}
\pagestyle{plain}
\fi
\maketitle

 \thispagestyle{fancy}

\begin{abstract}
Recent developments in deep generative models have opened up a wide range of opportunities for image synthesis, leading to significant changes in various creative fields, including the fashion industry. While numerous methods have been proposed to benefit buyers, particularly in virtual try-on applications, there has been relatively less focus on facilitating fast prototyping for designers and customers seeking to order new designs. To address this gap, we introduce DiCTI (Diffusion-based Clothing Designer via Text-guided Input), a straightforward yet highly effective approach that allows designers to quickly visualize fashion-related ideas using text inputs only. 
Given an image of a person and a description of the desired garments as input, DiCTI automatically generates multiple high-resolution, photorealistic images that capture the expressed semantics.  
By leveraging a powerful diffusion-based inpainting model conditioned on text inputs, DiCTI is able to synthesize convincing, high-quality images with varied clothing designs that viably follow the provided text descriptions, while being able to process very diverse and challenging inputs, captured in completely unconstrained settings. We evaluate DiCTI in comprehensive experiments on two different datasets (VITON-HD and Fashionpedia) and in comparison to the state-of-the-art (SoTa). The results of our experiments show that DiCTI convincingly outperforms the SoTA competitor in generating higher quality images with more elaborate garments and superior text prompt adherence, both according to standard quantitative evaluation measures and human ratings, generated as part of a user study. The source code of DiCTI will be made publicly available. 
\end{abstract}

\section{Introduction}

The fashion industry is a thriving billion-dollar business that engages a diverse array of stakeholders, ranging from manufacturers, retailers, and merchandisers to buyers and models \cite{cheng2021fashion}. Among these contributors, fashion designers play a pivotal role by leveraging their creative talents to craft innovative garment outfits that aim to resonate with the market and ultimately satisfy the discerning tastes of buyers. As the fashion industry continues to evolve, designers and consumers are adapting to these transformative trends. Designers are leveraging technology to bring their creations to life, while buyers are embracing the freedom to express their individuality through fashion \cite{plesh2023glassesgan,fele2022cvton,pernuvs2023maskfacegan}. 

Recently, generative models for image synthesis as well as text understanding models have undergone rapid advances~\cite{cao2024controllable}. State-of-the-art generative models, for example, are today capable of 
generating
a 
wide variety of realistic, high-resolution images of various scenes and objects~\cite{wang2018pix2pixHD,rombach2021highresolution,ramesh2021zeroshot}. 
The fashion industry also embraced generative technology to streamline different automation tasks, such as virtual try-on~\cite{plesh2023glassesgan,fele2022cvton,chou2018pivton} or make-up transfer~\cite{jiang2020psgan,zu2022tsevgan}.
\begin{figure}[t]
\includegraphics[width=0.95\columnwidth]{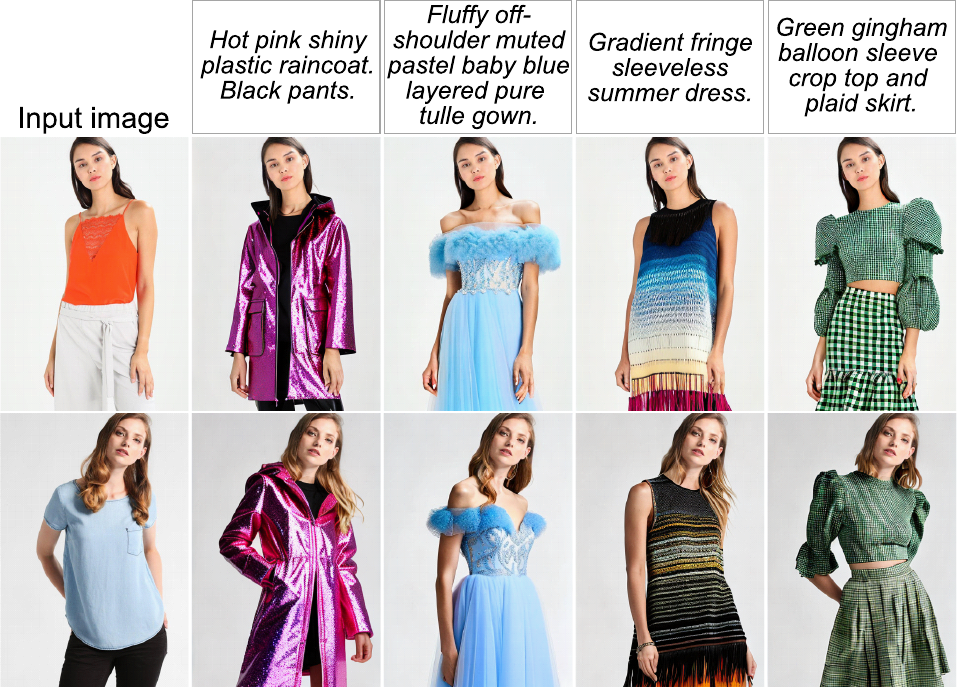}
\caption{\textbf{Example results generated by DiCTI.} 
Given an initial image and a description of the desired outfit, \modelname, the proposed model for text-guided garment design, produces a photo-realistic image with the person in the original image in an outfit that matches the provided text description.\vspace{-2mm}
}  
\label{fig:banner}
\end{figure}
Furthermore, researchers have also been looking into other interesting applications of computer vision models in the fashion industry, such as garment attribute transfer~\cite{Cui2021DressingIO} and editing~\cite{dadfar2023TDGEM}, outfit fashionability improvement~\cite{han2019finet,hsiao2019fashionplus} and controllable garment synthesis~\cite{Han2020FromDD,zhao2022DesignerGAN,pernus2023fice}. The recent success of text and vision models opened the door for advances in text-controlled fashion image synthesis~\cite{jiang2022text2human,pernus2023fice,baldrati2023multimodal}. Despite the similarities of these tasks, the target audience of different methods varies. While some explicitly target consumers (e.g., virtual try-on), others aim to assist designers in brainstorming ideas for new designs by allowing fine-grained spatial control over the generated results by providing additional inputs, such as sketches~\cite{baldrati2023multimodal,zhao2022DesignerGAN}. Different from such techniques, our goal in this work is to design a tool that supports consumers in communicating their wishes to a designer or allow for simple image search on the internet to find similar garments available in online stores. As illustrated in Fig. \ref{fig:banner}, we address this task with a text-conditioned image editing model capable of generating high-quality images of a provided input subject with a desired garment design. 

Specifically, we propose a novel approach to fashion image editing, termed DiCTI -- a \textbf{Di}ffusion-based
\textbf{C}lothing Designer via \textbf{T}ext-guided \textbf{I}nput, that requires solely an input image of a person and a text description of the desired garment expressed in natural language to generate creative and fashionable garments. With DiCTI, we formulate the image-editing task as an inpainintg problem, and leverage the expressivity of a pre-trained general purpose latent diffusion model~\cite{rombach2021highresolution} to generate visually convincing, varied and realistic garment designs based on the provided descriptions. 
We assess the performance of DiCTI on two diverse test datasets, i.e., VITON-HD~\cite{choi2021viton} and Fashionpedia~\cite{jia2020fashionpedia}, demonstrating that the approach is applicable to images captured in constrained, but also completely unconstrained (i.e., \textit{in-the-wild}) settings.  To validate the performance of DiCTI and put the reported results into context, we compare our method to the most similar work from the literature in terms of input requirements and goal, FICE~\cite{pernus2023fice}. Through comprehensive experiments, including a user study,  
we show that DiCTI can produce realistic images of clothed humans while successfully adhering to the manually prepared text prompts and does so significantly better than FICE.

\section{Related Work}

Controlled garment image synthesis and editing has recently become an active topic of research. A considerable amount of techniques has been proposed in the literature to guide the synthesis/editing process in recent years, including techniques that rely on \textit{spatial information} (e.g., pose representations, keypoints, sketches, etc.) and \textit{text descriptions}. Below, we briefly review some of these techniques to provide the necessary background for the proposed DiCTI model. 


\subsection{Guidance with spatial information}

Some methods generate pose or parsing maps of an input image to guide garment image synthesis. 
FiNet~\cite{han2019finet}, for example, aims to inpaint a missing piece of garment in the input image by first predicting its segmentation map and then generating a visually compatible garment that fits into the original image. However, this method gives the user little control over the synthesized garment. 
PISE~\cite{zhang2021PISE} similarly generates target parsing map based on pose keypoints and source parsing map, then uses it to guide synthesis of an image of a person in a different pose or with garment texture from another image. 
ADGAN~\cite{men2020controllable} learns image mapping into two latent spaces - a pose representation and a style representation that consists of disentangled components for different semantic regions in the image. This allows for a controlled image composition based on multiple image inputs, where style is independent from pose. 
Other authors provide design sketches to allow for better control over garment appearance.
D2RNET~\cite{Han2020FromDD}, for instance, consists of a two-branch pipeline that synthesizes a preview of real fashion items that correspond to the given design drafts. 

\subsection{Guidance with text}

With the advancements in text parsing, text understanding and successful association with image synthesis techniques (through models, such as CLIP~\cite{radford2021learning}), text prompts have become a popular way to guide the synthesis process in conjunction with other guiding parameters.
Text2Human~\cite{jiang2022text2human}, for example, proposes a two-step pipeline for synthesis of clothed humans. First, a semantic parsing map of the target image is predicted and then used together with a text prompt to guide image synthesis.
TD-GEM~\cite{dadfar2023TDGEM} learns to control partially disentangled style space latent vectors using text prompts, thus allowing for text-guided editing of garments. 
FICE~\cite{pernus2023fice} employs an extended GAN inversion technique to maximize the correspondence between the generated garment and the textual description, while preserving the initial identity and pose. 
The Multimodal Garment Designer~\cite{baldrati2023multimodal} incorporates text inputs together with sketch and pose conditioning to guide garment synthesis in the underlying denoising diffusion model. 
In our work, we relate to FICE and Multimodal Garment Designer, seeking the best of both worlds - simplicity of use of the former, requiring no additional inputs, and image quality and model expressivity on par with that of the latter.

\section{Methodology}
In this section, we present our proposed denoising diffusion-based method for text-guided garment synthesis, termed DiCTI, capable of generating highly realistic results, as illustrated in Fig. \ref{fig:banner}. We start the section with a formal description of the problem setting and then describe the individual components of the proposed editing approach.

\subsection{Problem formulation and DiCTI overview}
Given a reference image of a person $I$ and a text description of the desired garment $y$, the goal of DiCTI is to generate an image $\hat{I}$ of the same person with the person's garments substituted for garments corresponding to the text description. 
To achieve this goal, we formulate the editing problem as an inpainting task, where the generative model is given a mask of the initial clothing area, and is tasked with filling it in, while keeping the rest of the image unaltered. 

\begin{figure*}
\centering
\includegraphics[width=0.97\textwidth]{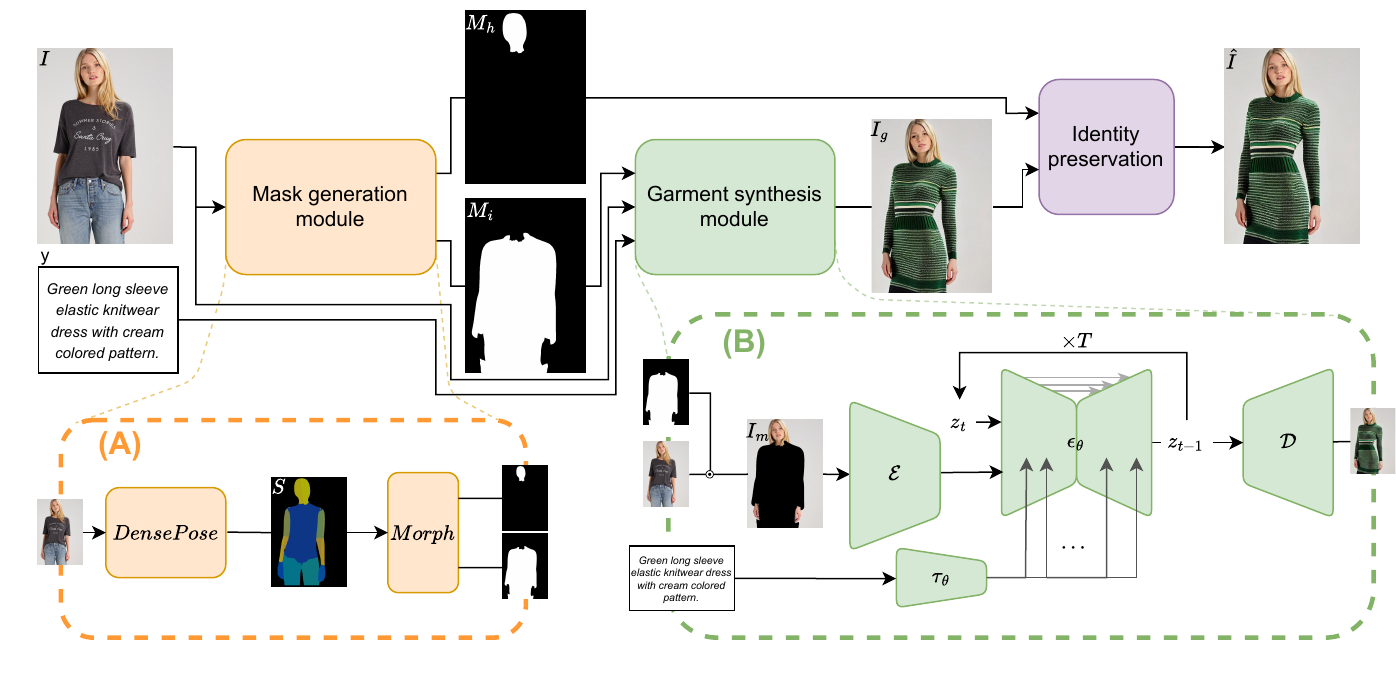}
\caption{
\textbf{High-level overview of the proposed DiCTI method.} DiCTI consists of multiple components. The \textit{Mask Generation Module} (A) generates a binary mask covering the body/clothing and another one covering the head of the person in the input image. The body mask, along with the input image, is then passed to the \textit{Garment Synthesis Module} (B), responsible for completing the masked-out parts of the image in adherence to the prompt. The synthesized image then undergoes post-processing to restore facial features that may have been altered during synthesis and ensure \textit{Identity Preservation}.}
\vspace{-2ex}
\label{fig:pipeline_overview}
\end{figure*}

Fig. \ref{fig:pipeline_overview} shows a high-level overview of the proposed method. As can be seen, the input to DiCTI is an image and a text prompt with the description of the desired clothing. The image is first passed through the \textit{Mask Generation Module} (\secref{Sec: MAsks}) to infer the area that needs to be inpainted as well as the position of the face. The resulting body mask, together with the rest of the inputs, is then passed through the \textit{Garment Synthesis Module} (\secref{Sec:Inapint}), which generates new garments in the masked area according to the text description. Finally, an additional \textit{Identity Preservation} (\secref{Sec:Postpr}) stage is utilized to ensure the preservation of facial features.

\subsection{Mask Generation Module\label{Sec: MAsks}}
The first component of DiCTI is the Mask Generation Module (MGM). It takes an input image and generates two binary masks: $(i)$ an inpainting mask $M_i$, covering the entirety of the human body, except for the face, hands, and feet, and $(ii)$ a head mask $M_h$ for the post-processing (identity preservation) step. 
In order to produce such masks, the relevant regions in the input image need to be identified. With the goal of preventing the initial garment shape from influencing the shape of the synthesized garment, we base the MGM on a clothing-agnostic parser. Specifically, we use DensePose~\cite{guler2018densepose} to generate a map of labels for the input image. DensePose is a human-pose parsing model, which has shown great robustness to a variety of poses as well as background distractors \cite{jug2022body}. When applying DensePose, each image pixel in $I$ is assigned to one of the $24$ body part categories or the background: $S^{h \times w}$, where $S(i, j) \in [0, 24]$.

\vspace{1.5mm}\noindent\textbf{Computing the inpainting mask.} The inpainting mask $M_i$ is generated from the DensePose output $S$ and through the application of various morphological operations. To compute the inpainting mask, we first combine the binary masks of all body parts, except for the face, hands and feet, and produce a body-part mask $M_b$. To account for loose-fitting clothing, the mask is dilated by $d$ pixels in every direction. Next, we combine areas that need to preserved (corresponding to the face, hands, and feet) into a preservation-area mask $M_p$ and erode the mask by $e$ pixels in each direction. Finally, we subtract the eroded preservation-area mask form the dilated body mask. This results in an inpainting mask that covers all existing clothing, allows freedom in generating new garments, and prevents the identity of the person in $I$ from being altered. Formally, this process is defined as follows: 
\begin{equation}
M_i = M_b \oplus C^d - (M_p \ominus C^e),
\end{equation}
where $C^x$ is a circle-shaped structuring element of radius $x$, and $\oplus$ and $\ominus$ are morphological dilation and erosion operations, respectively.

\vspace{1.5mm}\noindent\textbf{Computing the head mask.} To ensure identity preservation, which is especially noticeable in facial features, we perform image stitching in the post-processing phase. A mask is needed to determine the location of the face and neck within the original image. Similarly as with the inpainting mask, the DensePose segmentation $S$ is used to determine the area of interest, which is then eroded by $f$ pixels to avoid sharp edges and artifacts in case of small misalignments, i.e.: 
\begin{equation}
M_h = M_{he} \ominus C^f,
\end{equation}
where $M_{he}$ is a binary mask corresponding to the head region in $S$ and $C^f$ is again a circle-shaped structuring element.    


\subsection{Garment Synthesis Module\label{Sec:Inapint}}
The second component of DiCTI is the Garment Synthesis Module (GSM), which inpaints the area covered by $M_i$ in $I$ in accordance with the provided text prompt $y$. For the synthesis module, we leverage the generalization abilities of a latent diffusion-based model (i.e., Stable-Diffusion-V2\footnote{Available from: {\scriptsize \url{https://huggingface.co/stabilityai/stable-diffusion-2-inpainting}}}) pretrained  on a large amount of data. We choose not to finetune the pretrained model to avoid a loss in generality, which is inevitable especially in the case of smaller datasets, such as those generally used for clothed-human synthesis. This way, the model may be tasked with generating rather uncommon outfits, that may not be seen in specialized datasets. Thus, general understanding of different concepts is more desirable than narrow specialization for the task at hand. To make the paper self-contained, we first present the background on denoising diffusion models and then discuss their application for image inpainting in our GSM.

\vspace{1.5mm}\noindent\textbf{Prerequisites on denoising diffusion models.} Denoising diffusion models~\cite{Ho2020DenoisingDP} are based on the assumption that progressively adding Gaussian noise to a clear input image $x_0$ will result in an all-noise image after $T$ steps. 
The idea is to sample $x_t$ at timesteps $t \in \left [ 1, T \right ]$ given the closed-form formula
\begin{equation}
x_t \sim \mathcal{N}(x_t; \sqrt{\bar{\alpha_t}}x_0, (1-\bar{\alpha}_t) \mathbf{I}),
\end{equation}
where $\bar{\alpha_t} = \Pi_{s=0}^t \alpha s$, $\alpha = 1 - \beta_t$ and $\beta_t$ is the variance of the Gaussian noise added to $x_{t-1}$, and then use the corresponding pairs of data points $x_i$, $x_{i-1}$ to learn the reverse (i.e., backward diffusion) process, i.e., to approximate $q(x_{t-1}|x_t)$. Calculating for all $t$ from $T$ to $0$ then yields a sample from the data distribution
\begin{equation*}
p_\theta(x_{0:T}) = p_\theta(x_T).
\end{equation*}
To reduce the high computational cost of iteratively performing reverse diffusion on images in pixel space, Latent Diffusion Models (LDM)~\cite{rombach2021highresolution} reduce the dimensionality of the problem by projecting the image onto a latent space of a pre-trained encoder $\mathcal{E}$. Additionally, they introduce model conditioning by jointly optimizing the model $\epsilon_\theta$ and a domain-specific expert model $\tau_\theta$ whose outputs are passed to the model using a cross-attention mechanism. The learning objective then becomes
\begin{equation}
L_\text{LDM} = \mathbb{E}_{\mathcal{E}(x),y,t,\epsilon \sim \mathcal{N}(0, I)} \left [ \left \| \epsilon - \epsilon_\theta(z_t, t, \tau_\theta(y) \right \| ^2 \right ],
\end{equation}
where $z_t$ is a latent representation of $x_t$ and $y$ is a conditioning input, such as a text prompt or a semantic map. 

\vspace{1.5mm}\noindent\textbf{Inpainting Latent Diffusion Models}~\cite{rombach2021highresolution} follow the same objective as the LDM models presented above, but limit the generative process within the backward diffusion steps to the area determined by a mask. From an implementational point of view, the pretrained backbone diffusion model receives five additional input channels to accommodate the masked image and inpainting mask.
In addition to the latent vector $z_t \in \mathbb{R}^{w\times h \times 4}$, time step $t$ and text prompt $y$, the model receives the inpainting mask ($M_i \in \mathbb{R}^{W \times H \times 1}$) and a masked image $I_m = (1 - M_i) \odot I \in \mathbb{R}^{W \times H \times 3}$, containing only areas that should be preserved. The masked image is encoded with encoder $\mathcal{E}$ to reduce its dimensionality to $\mathcal{E}(I_m) \in \mathbb{R}^{w\times h \times 4}$ where $h < H$ and $w < W$. The inpainting mask is downsampled to match the latent dimensions $w$ and $h$. The latent vector, inpainting mask, and masked image are concatenated along the channel dimension and passed to the diffusion model that outputs prediction for the noise at time step $t-1$. The denoising process is further guided by the text prompt, enhanced with some additional words, such as {\small \texttt{photorealism}, \texttt{detailed hands}, \texttt{natural lightning}}, and {\small\texttt{sharp}} to encourage these traits in an image. 
Text embeddings 
are computed using a pre-trained OpenCLIP~\cite{ilharco2021openclip} model ($\tau_\theta(y)$). The result of the inpainting latent diffusion is an intermediate image $I_g$ with the synthesized clothing. However, since the images often contain faces that differ in appearance from the faces in the input image $I$, we use a final post-processing step to ensure better identity preservation.

\subsection{Identity preservation\label{Sec:Postpr}}
Due to the inherent weakness and imperfections of generative models, preserving the appearance and facial features of the subjects in the input images is challenging and such features are often changed and smoothed during the synthesis process. To better preserve facial appearance, we perform an image stitching operation in the post-processing stage, i.e.:
\begin{equation}
\hat{I} = I_g \odot (1 - M_h) + I \odot M_h,
\end{equation}
where $I_g$ is the output of the generative model, $I$ is the source image and $M_h$ is the binary head mask. This operation is possible since the generator  model preserves the pose of the original image around the neck area. The output of the presented post-processing step is the final edited image $\hat{I}$.


\section{Experiments and Results}
In this section, we first describe our experimental setting as well as the  datasets and quantitative performance measures used for the evaluation \cite{krivzaj2024deep}. 
Due to the difficulty of capturing the performance of generative methods without a unique ground truth in a single meaningful number, we perform a series of experiments that allow us to illustrate the strong and weak points of DiCTI and compare it to a competing method.  
We start the experimental section with a comparison to a SoTa competitor, i.e., FICE~\cite{pernus2023fice}, to put the capabilities if DiCTI into perspective. To this end, we first report a number of image quality scores that measure image realism and adherence to the text prompt as perceived by deep neural models. However, compared to computers, humans put more emphasis on different image characteristics and, as a result, automated evaluation techniques often do not correlate entirely with human perception. Thus, we perform a user study to gauge human perception of the performance of DiCTI compared to FICE. 
Next, we present a qualitative evaluation and conduct an ablation study to determine the effect of the mask-dilation parameter $d$ on the results of synthesis. We conclude the section with a visual analysis of the generated images, to explore the limitations of our approach and to identify directions for further research.

\subsection{Datasets}
\label{sec:eval/datasets}
We test DiCTI on two challenging high-resolution fashion datasets, i.e., VITON-HD~\cite{choi2021viton} and Fashionpedia~\cite{jia2020fashionpedia}. VITON-HD is a collection of 13,679 commercial images of models in a frontal pose, and their corresponding upper-garment. The images are taken in a controlled environment with posing models in front of a homogeneous light colored background and with the depicted person taking up the larger part of the image. Most of the models are only visible from knees or middle of the thighs up. Fashionpedia~\cite{jia2020fashionpedia}, on the other hand, is a set of 48,825 fashion images of celebrities captured in daily life, i.e., in the wild. It includes images of various sizes and aspect ratios, covering a number of challenging scenarios, such as background clutter, imperfect lighting conditions, non-target human distractors, and uncommon and challenging poses.
We therefore use VITON-HD to evaluate performance in a controlled environment and establish performance upper-bounds of the method, whereas the purpose of Fashionpedia is to measure robustness when dealing with photos taken \textit{in the wild}.

\begin{figure}[t]
\includegraphics[width=0.97\columnwidth]{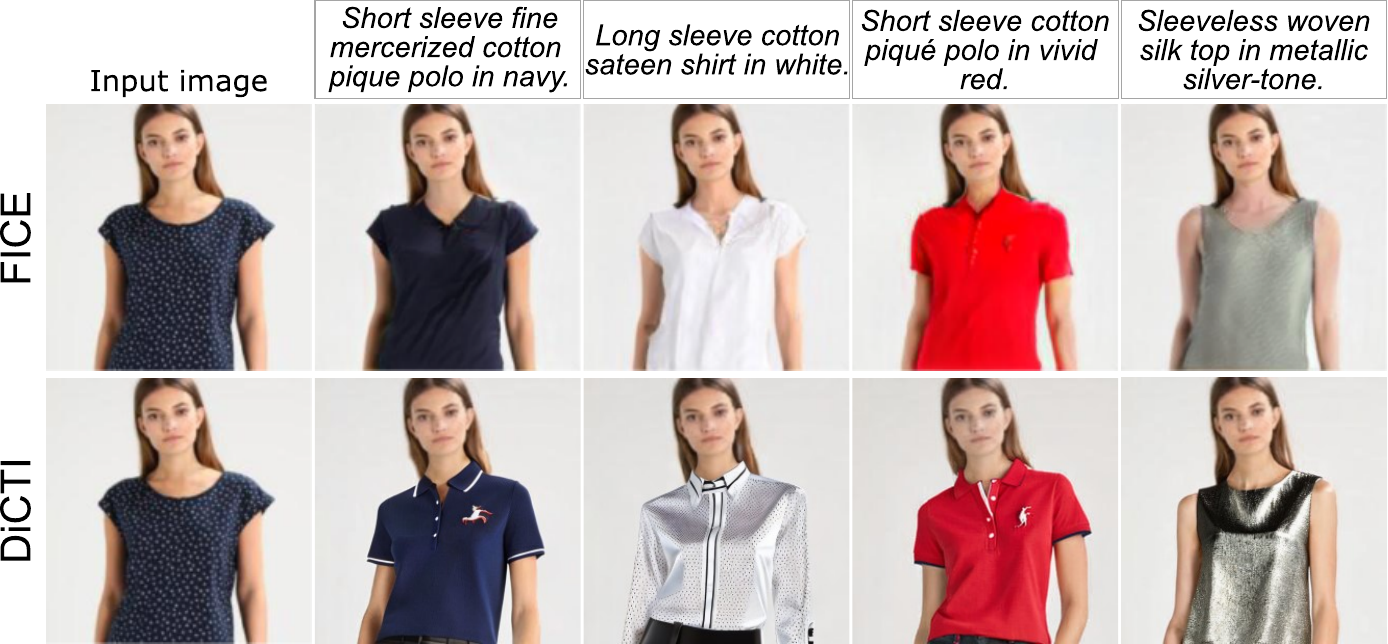}
\caption{\textbf{Comparison of FICE and DiCTI.} Example synthesis results are presented for various text prompts.\vspace{-1mm}}
\label{fig:fice_examples}
\end{figure}

\begin{table}[t]
\caption{\textbf{Qualitative comparison of FICE and DiCTI.} KID, CLIP-S and CLIP-IQA scores are reported.\vspace{-1mm}}
\label{tab:quantitative}
\center
\begin{tabular}{l | c c c}
\toprule
\textbf{Method} & \textbf{KID} $\downarrow$ & \textbf{CLIP-S} $\uparrow$ & \textbf{CLIP-IQA} $\uparrow$ \\
\midrule
FICE~\cite{pernus2023fice} & $0.082$ & $22.04$ & $0.58$ \\
DiCTI (ours) & $\mathbf{0.066}$ & $\mathbf{27.48}$ & $\mathbf{0.72}$ \\
\bottomrule
\end{tabular}\vspace{-2mm}
\end{table}

\subsection{Performance measures}
\label{sec:measures}
Since evaluating image synthesis techniques quantitatively is a complex task, we use a collection of image quality assessment (IQA) measures that capture different image characteristics. We evaluate visual image quality with the Kernel Inception Distance score (KID)~\cite{binkowski2018demystifying}, which compares distributions of real and generated images. Specifically, we use the implementation proposed by~\cite{parmar2021cleanfid}. Since KID measures the distance between two distributions, a lower value is preferred, meaning that the distribution of generated images is closer to that of the real images, implying that the images look more realistic. 
Additionally, we employ the CLIP-IQA~\cite{wang2022exploring} quality score that uses a pair of antonyms as an input (e.g. \texttt{["Good photo.", "Bad photo."]}) and considers softmax values of the cosine similarity of the image embedding to each of the prompts:
\begin{equation*}
\text{CLIP-IQA} = \frac{e^{s_{pos}}}{e^{s_{pos}} + e^{s_{neg}}},
\end{equation*}
where $s_{pos}$ is the cosine similarity between CLIP embeddings of the input image $I$ and the positive description (e. g. \texttt{Good photo.}) and $s_{neg}$ is cosine similarity between the input image and the negative description (e. g. \texttt{Bad photo.}). The value of CLIP-IQA is between 0 and 1, where higher value indicates a better match with the positive description. For the description pair above, the higher score indicates a picture of a higher quality. In addition to realistic appearance, a successful method for the task of garment synthesis should also ensure adherence to the text prompt. We use the CLIP score~\cite{hessel2021clipscore} (CLIP-S) to measure the similarity between the input image ($I$) and the text prompt ($y$) CLIP embeddings $c_I$ and $c_y$:
\begin{equation*}
\text{CLIP-S}(c_I, c_y) = 100 \cdot \max(\cos(c_I, c_y), 0).
\end{equation*}
The CLIP score range is $[0,100]$, where higher values indicate better image correspondence with the prompt $y$.

\begin{table}[t]
\caption{\textbf{Human-study results.} The reported statistics are aggregated per image for each scoring criterion, showing a fraction of times DiCTI was chosen over FICE.}
\label{tab:user_survey}
\centering
\begin{tabular}{l|c c c c}
\toprule
\textbf{Criterion} & \textbf{Mean} & \textbf{Std} & \textbf{Min} & \textbf{Max} \\
\midrule
Identity & $0.67$ & $0.16$ & $0.41$ & $0.92$ \\
Pose & $0.22$ & $0.17$ & $0.00$ & $0.62$ \\
Prompt & $0.88$ & $0.10$ & $0.67$ & $0.97$ \\
Realism & $0.72$ & $0.12$ & $0.45$ & $0.89$ \\
\bottomrule
\end{tabular}\vspace{-2mm}

\end{table}

\subsection{Comparison with the state-of-the-art}
\label{sec:eval/comparison}

\noindent\textbf{Quantitative comparison.} We compare our method to FICE~\cite{pernus2023fice}, a  SoTa method that operates on the same inputs -- an input image and a text prompt written in natural language. Unlike our method, FICE works on an image resolution of $256 \times 256$. Since the training of FICE was done on VITON-HD~\cite{choi2021viton}, the method is limited to female models and upper garments only. Therefore, the comparison is done on images from the VITON-HD test set, cropped to a square covering the upper part of the image, to ensure a fair comparison. The test set consists of $416$ images. In this experiment, we pair each of them with each of the $9$ pre-prepared prompts, to generate $3744$ images per method. 

Fig.~\ref{fig:fice_examples} shows the results produced by our method and FICE on the same example images and prompts as used in the original paper \cite{pernus2023fice}. As can be seen, DiCTI tends to generate much more detailed images with logos and sharp, accurate necklines (e.g. polo). This can, in part, be attributed to the higher-resolution generative model used in our method, which is also more general (with superior zero-shot capabilities), since it was never finetuned on a particular dataset. Furthermore, it partially stems from the tight-fitting masks used in the optimization-based GAN inversion phase of FICE, which only cover up the clothing part of the image and do not account for potentially more loose-fitting garments. This likely results in FICE being discouraged from generating additional garment parts, such as a polo neck. 

    

To get a numerical comparison of DiCTI and FICE, we compare the two in terms of the performance measures, described in~\secref{sec:measures}. Table~\ref{tab:quantitative} summarizes the values of KID, CLIP-score and CLIP-IQA, showing that DiCTI outperforms FICE, both in terms of image realism and adherence to the text prompts by a considerable margin. In other words, it better captures the expressed semantics while producing higher-quality images with superior realism.

\begin{figure}[t]
\centering
\includegraphics[width=0.97\columnwidth]{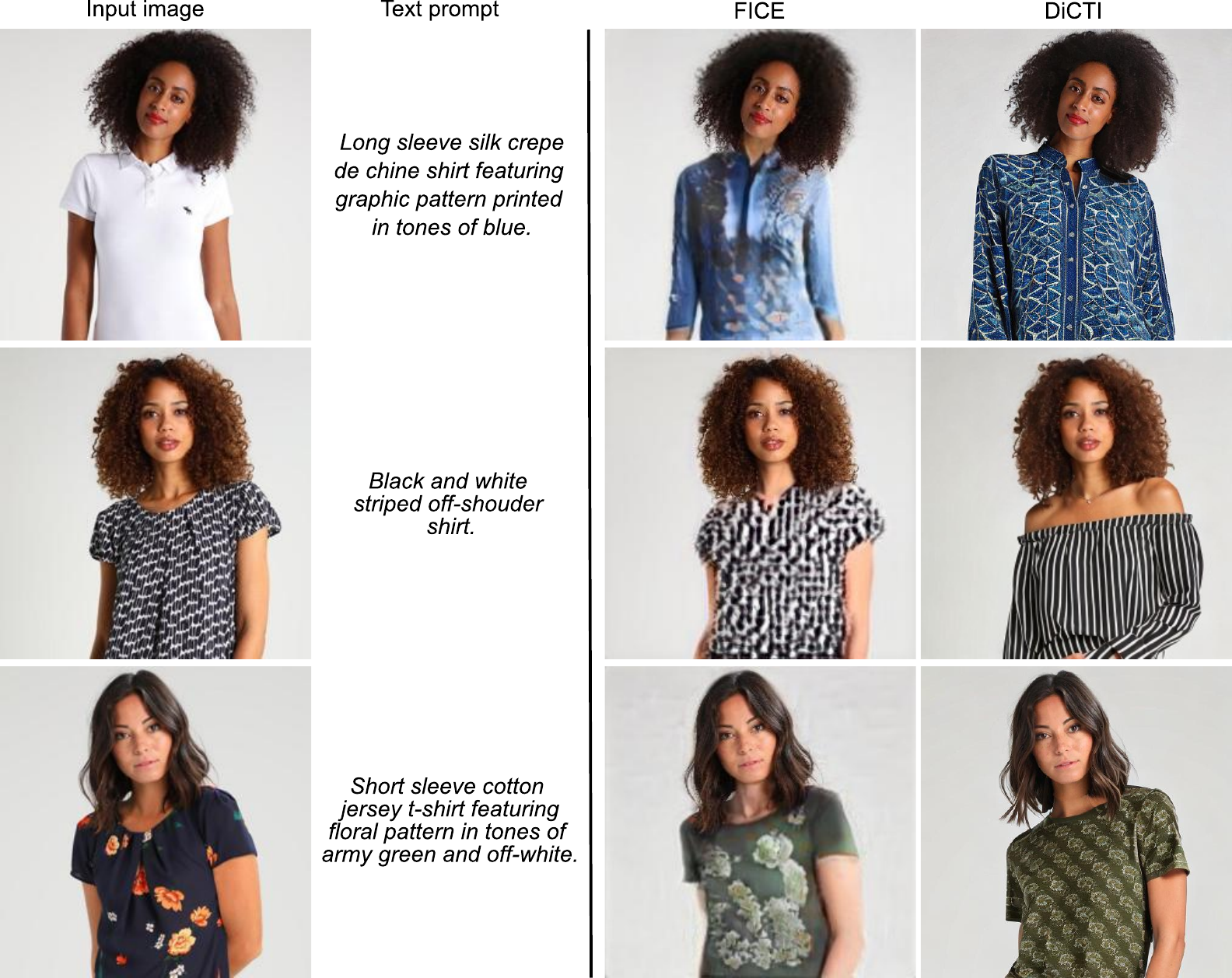}
\caption{\textbf{Examples from the user study.} While DiCTI occasionally alters the pose slightly, the results are commonly of higher quality and more faithful to the text prompt.\vspace{0mm}}
\label{fig:user-survey}
\end{figure}

\vspace{1mm}\noindent\textbf{Human/user study.} We validate the results of the reported quantitative evaluation through a human study. 
To this end, we uniformly sample two images from the VITON-HD test set per each of the nine text prompts. For each of the eighteen text-image pairs, we then ask participants to vote for the better performing model, i.e., DiCTI or FICE, w.r.t. $4$ criteria:
\begin{itemize}
    \item Q1: Perceived image realism,
    \item Q2: Adherence to the input text prompt,
    \item Q3: Preservation of the initial identity, and
    \item Q4: Preservation of initial pose.
\end{itemize}
The survey involved $39$ participants, yielding $702$ responses per question with image-text pairs, as shown in Fig. \ref{fig:user-survey}. 

\begin{figure}[t]
    \centering
    \includegraphics[width=0.97\columnwidth]{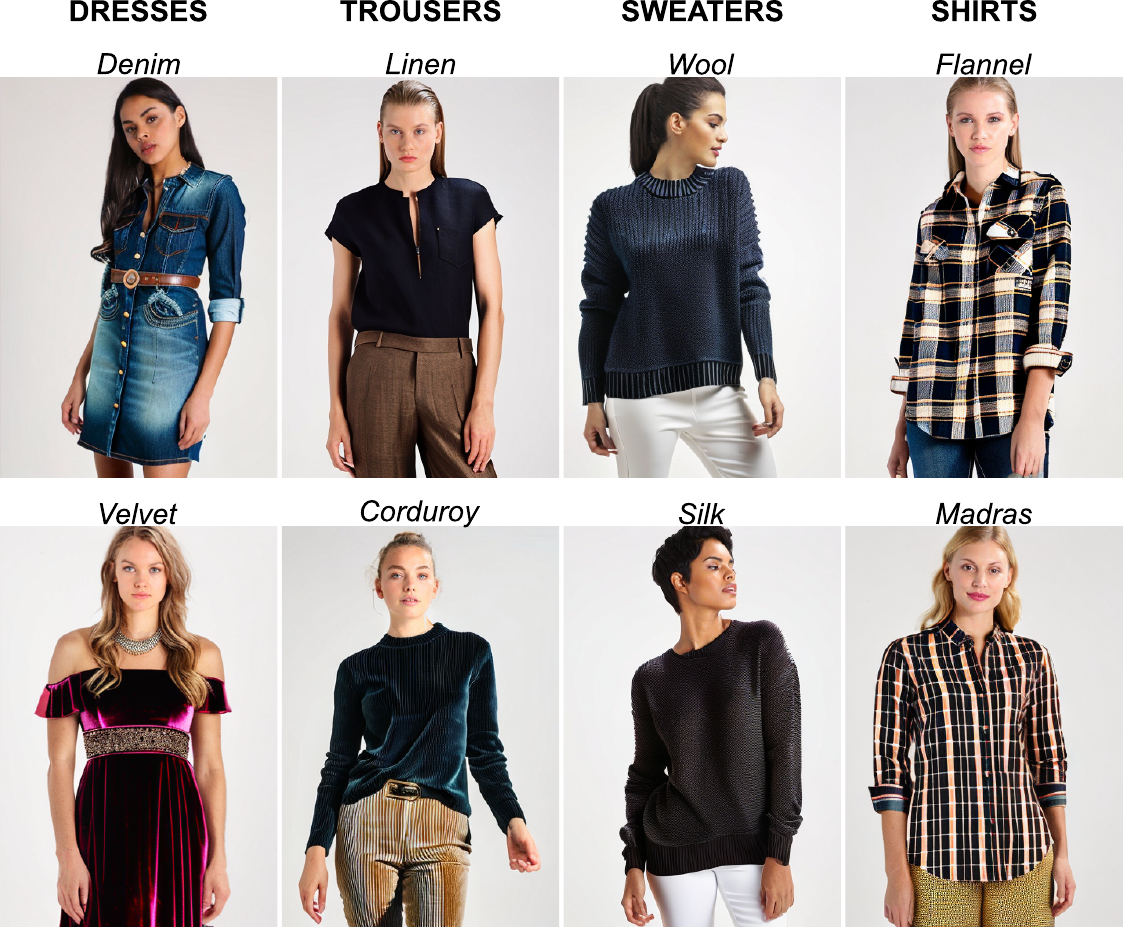}
    \caption{\textbf{Examples of different fabrics used for dresses, trousers, sweaters and shirts.} The prompts were generated by pairing garment property and type.\vspace{-2mm}}
    \label{fig:results_fabrics}
\end{figure}

\begin{figure}[t]
    \centering
    \includegraphics[width=0.97\columnwidth]{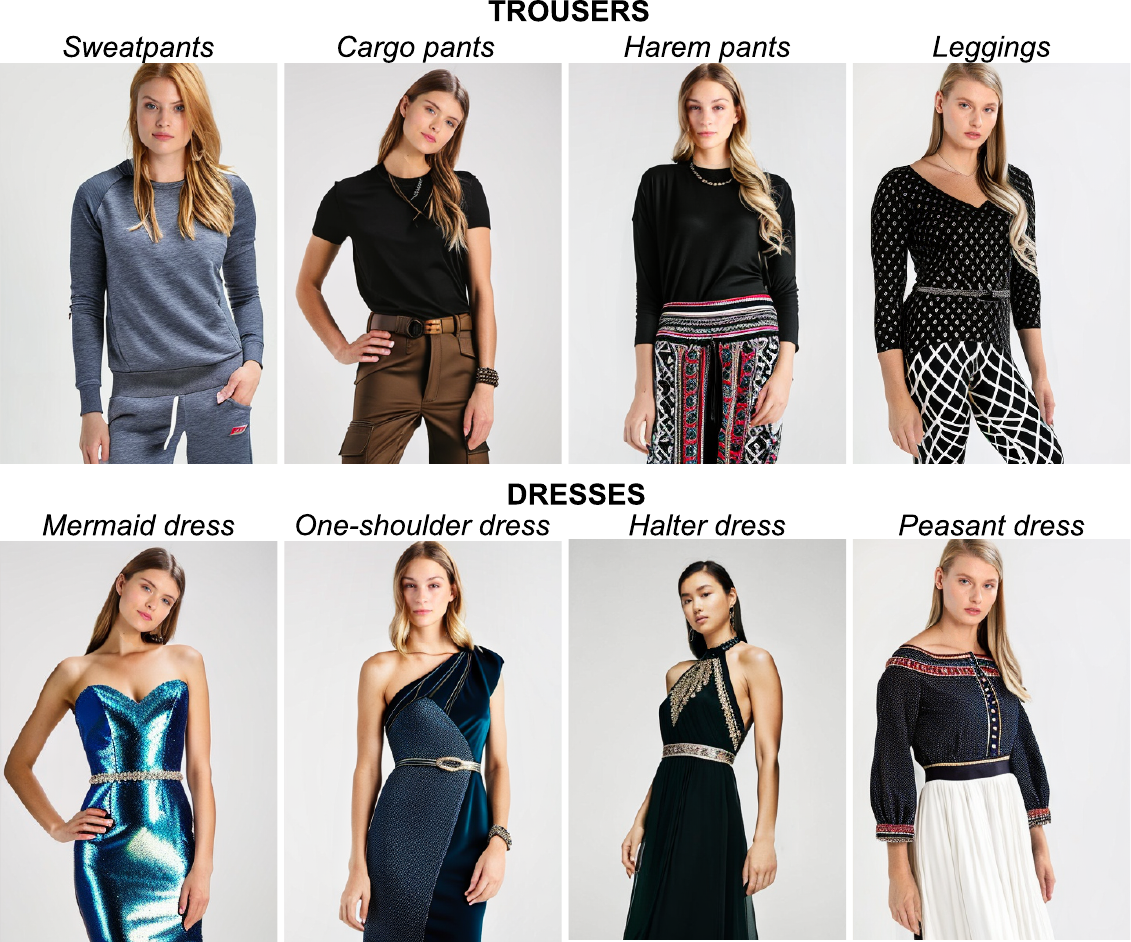}
    \caption{\textbf{Examples of different shapes used for trousers and dresses.} Note the realism, details and convincing textures generated based on the text prompts. In contract to FICE, DiCTI is also capable of generating lower garment results.\vspace{-2mm}}
    \label{fig:results_shapes}
\end{figure}

We aggregate the percentage of times that an image generated by DiCTI was chosen over one generated by FICE. The statistics are summarized in Tab.~\ref{tab:user_survey}. We perform a t-test, which confirmed that the results are statistically significant for all the scoring criteria. The results show that our method performs worse in terms of pose preservation, which is partially expected, given that DiCTI enforces no explicit conditioning on the model pose.
Due to the mask dilation applied in the mask generation phase, part of the information about the body pose gets lost. On the other hand, even when the pose does get altered, it is done without reducing the realism of the image. In contrast, FICE constrains the editing area to a tight mask covering an existing garment, which encourages pose preservation, however, at the cost of limiting the garment fit to that of the original clothing. 

DiCTI clearly outperforms FICE in the rest of the scoring criteria. When scoring \textit{identity preservation}, it is selected over FICE 67\% of the time, showing the effectiveness of the identity preservation module. 
Given the similarity of the identity preservation methodology of both methods,
we speculate the decisive factor to be the preservation of the skin tone. FICE often results in altered skin color compared to the original, while DiCTI generally synthesizes skin tone, consistent with the input image (rows two and three of Fig.~\ref{fig:user-survey}).
The difference in performance is even more apparent for the realism and text prompt adherence criteria. On average, DiCTI is selected over FICE 72\% and 88\% of the time for realism and prompt adherence, respectively, while the standard deviations for those criteria are lower than in the case of identity and pose preservation. This suggests that DiCTI is capable of leveraging the underlying pre-trained models' ability to generate high-quality, high-resolution images, while also adhering to the text prompt to a high degree. Conversely, 
FICE images tend to get blurry and less accurate in terms of the desired garment style and patterns.
\begin{figure}[t]
    \centering
    \includegraphics[width=0.97\columnwidth]{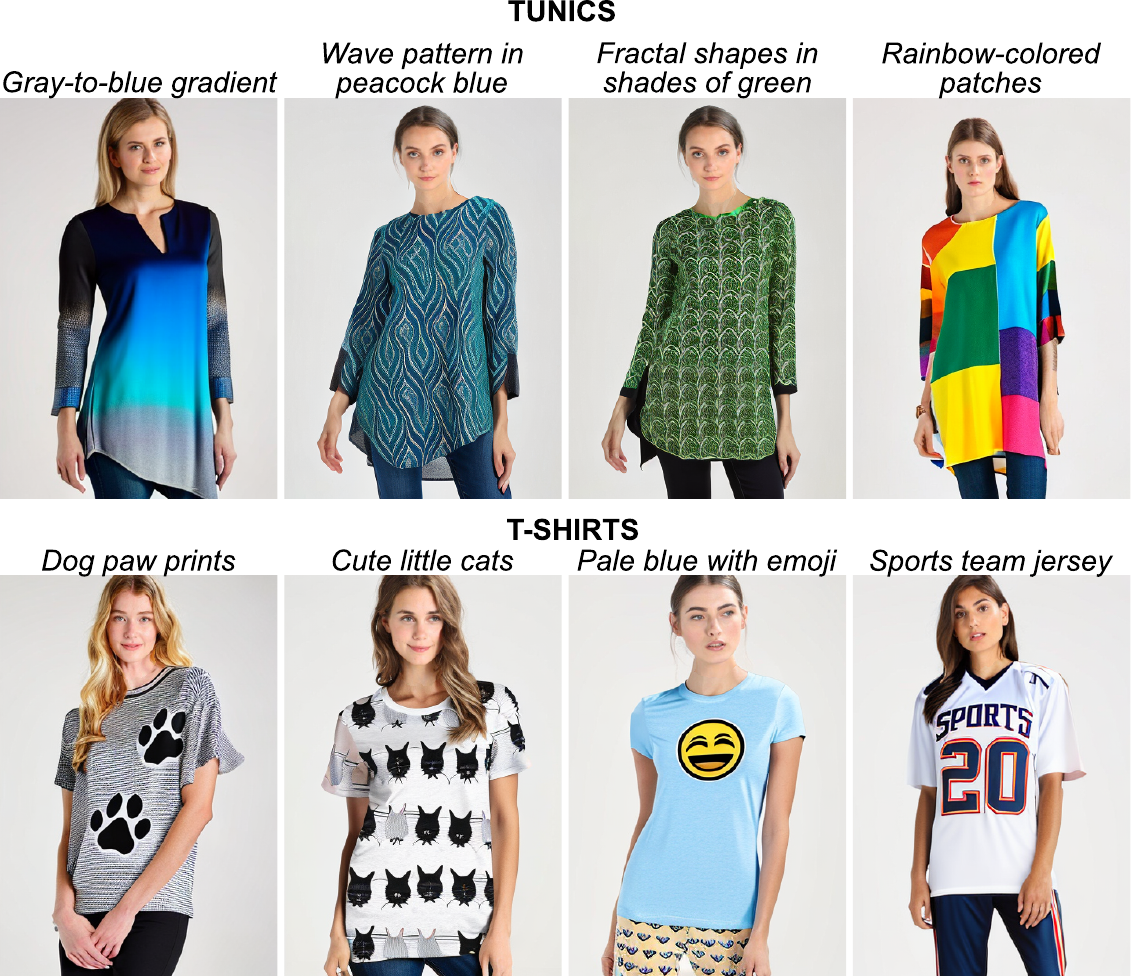}
    \caption{\textbf{Examples of different patterns and prints on tunics and T-shirts.} The prompts pair garment property and type.\vspace{-0mm}}
    \label{fig:results_patterns}
\end{figure}
\begin{figure}[!h!]
\centering
\includegraphics[width=0.97\columnwidth]{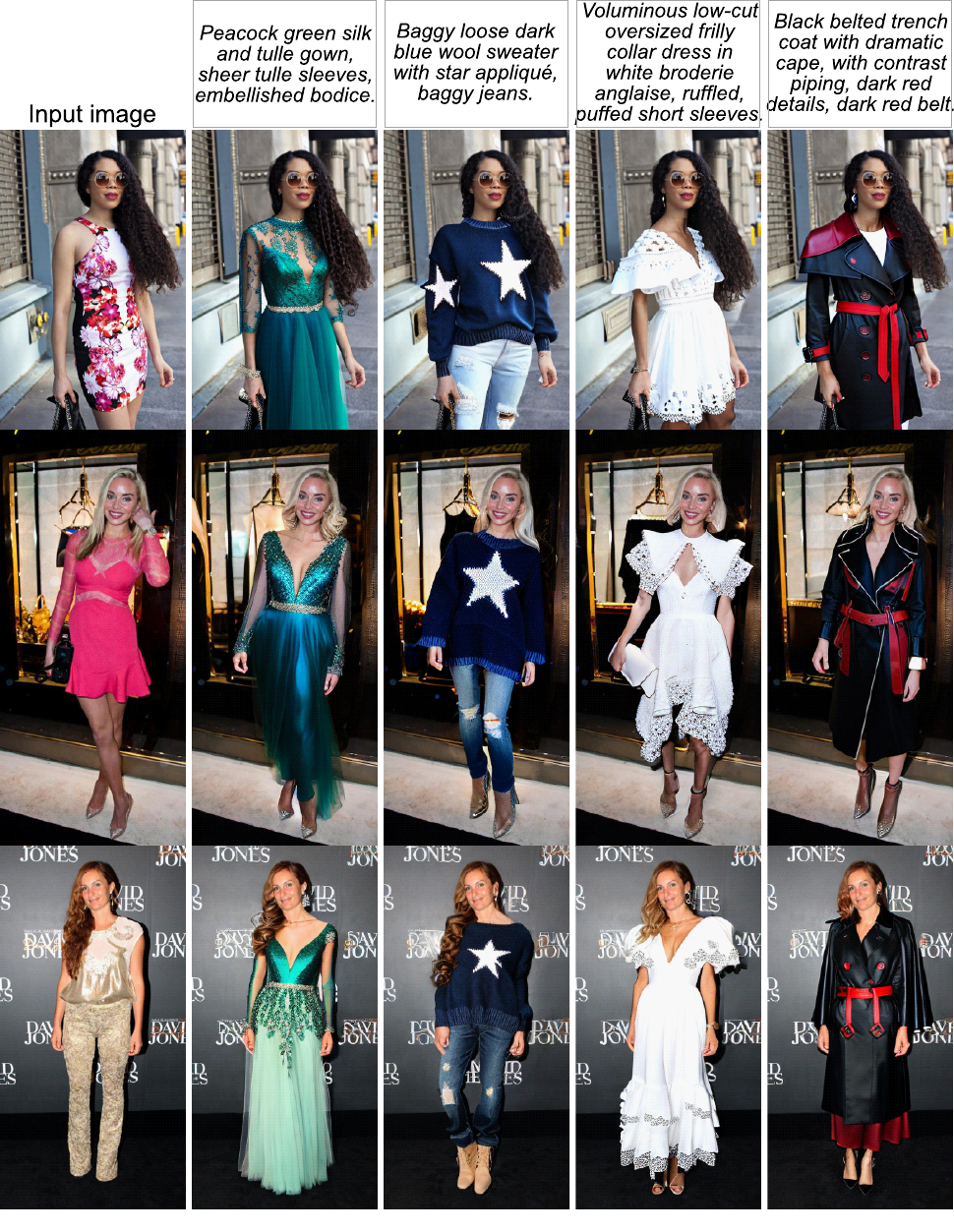}
\caption{\textbf{Visual examples with female models from Fashionpedia.} The first column shows the input images. Each of the subsequent columns shows results of one text prompt.\vspace{-3mm}}
\label{fig:results_fashionpedia_female}
\end{figure}

\begin{figure}[t]
\centering
\includegraphics[width=0.97\linewidth]{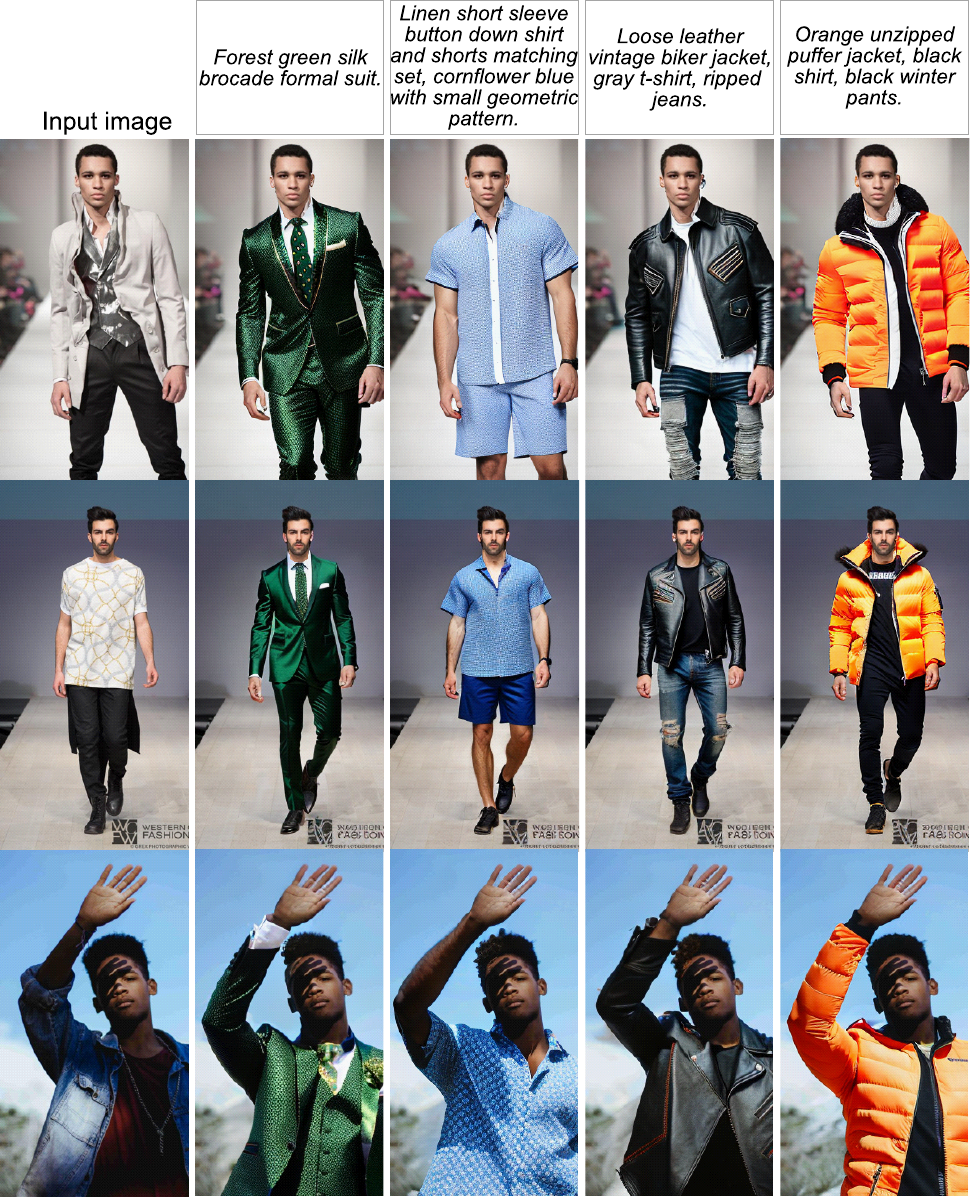}
\caption{\textbf{Visual examples with male models from Fashionpedia.} The first column shows the input images. Each of the subsequent columns shows results of one text prompt.\vspace{-3mm}}
\label{fig:results_fashionpedia_male}
\end{figure}
\subsection{Qualitative analysis}
\label{sec:eval/analysis}
Next, we analyze some of the synthesized images visually to gain better insights into DiCTI's strengths and weaknesses. 
For this, we test DiCTI on a number of text prompts describing different garment fabrics, shapes, and patterns.

We begin by isolating some of the garment features and letting the rest be freely chosen by the model. Fig.~\ref{fig:results_fabrics} shows that DiCTI can generate a number of different fabric types, such as silk, wool, velvet, or denim, on different kinds of garments. Fig.~\ref{fig:results_shapes} displays various garment shapes, according to length, sleeve and neck type, and fit. Fig.~\ref{fig:results_patterns} shows a number of different patterns and graphics to show DiCTI's versatility at understanding concepts deemed uncommon in the garment descriptions, such as cute little cats or emoji. 

Next, we focus on the Fashionpedia~\cite{jia2020fashionpedia} dataset, due to its diversity in gender and ethnicity of the subjects as well as the more realistic setting in which images have been taken. For this experiment, we use more verbose prompts, generally describing the whole outfits, while trying to cover as many garment features as possible.
Fig.~\ref{fig:results_fashionpedia_female} and Fig.~\ref{fig:results_fashionpedia_male} show some examples of images from the Fashionpedia~\cite{jia2020fashionpedia} dataset for female and male subjects, respectively. The results show that our method can perform well regardless of the gender and can also handle different ethnicities of the subjects depicted. Additionally, it highlights DiCTI's ability to understand the combinations of outfits. Furthermore, the results on the images from Fashionpedia show that the method is robust to background clutter and does not result in significant alterations of the background, with the occasional exception of some background text and image watermarks, which we argue are not critical given the goal of the task. 

\begin{table}[t]
    \centering
\caption{\textbf{Ablation study results.} KID, CLIP-S and CLIP-IQA scores are reported for different dilation sizes $d$.\vspace{-1mm}}
    \begin{tabular}{ l | c c c | c c c}
        \toprule
        & \multicolumn{3}{c}{\textbf{VITON}} & \multicolumn{3}{c}{\textbf{Fashionpedia}} \\ \cmidrule{2-4}\cmidrule{5-7}
        ${d}$ & \textbf{KID} $\downarrow$ & \textbf{CLIP-S} $\uparrow$ & \textbf{CLIP-IQA} $\uparrow$  & \textbf{KID} $\downarrow$ & \textbf{CLIP-S} $\uparrow$ & \textbf{CLIP-IQA} $\uparrow$\\
        \midrule
        $0$ & $\mathbf{0.057}$ & $24.52$ & $0.691$ & $\mathbf{0.010}$ & $19.63$ & $0.867$ \\
        $30$ & $0.068$ & $26.74$ & $0.688$ & $0.011$ & $21.5$ & $0.872$ \\
        $50$ & $0.071$ & $27.08$ & $0.698$ & $0.012$ & $22.03$ & $0.875$ \\
        $70$ & $0.073$ & $27.27$ & $\mathbf{0.700}$ & $0.013$ & $22.45$ & $0.876$ \\
        $90$ &$ 0.076$ & $27.32$ & $0.698$ & $0.014$ & $22.65$ & $\mathbf{0.881}$ \\
        $110$ & $0.077$ & $\mathbf{27.37}$ & $0.697$ & $0.016$ & $\mathbf{22.79}$ & $\mathbf{0.881}$ \\
        \bottomrule
    \end{tabular}
    \label{tab:ablation}
\end{table}

\begin{figure}[t]
    \centering
    \includegraphics[width=\columnwidth]{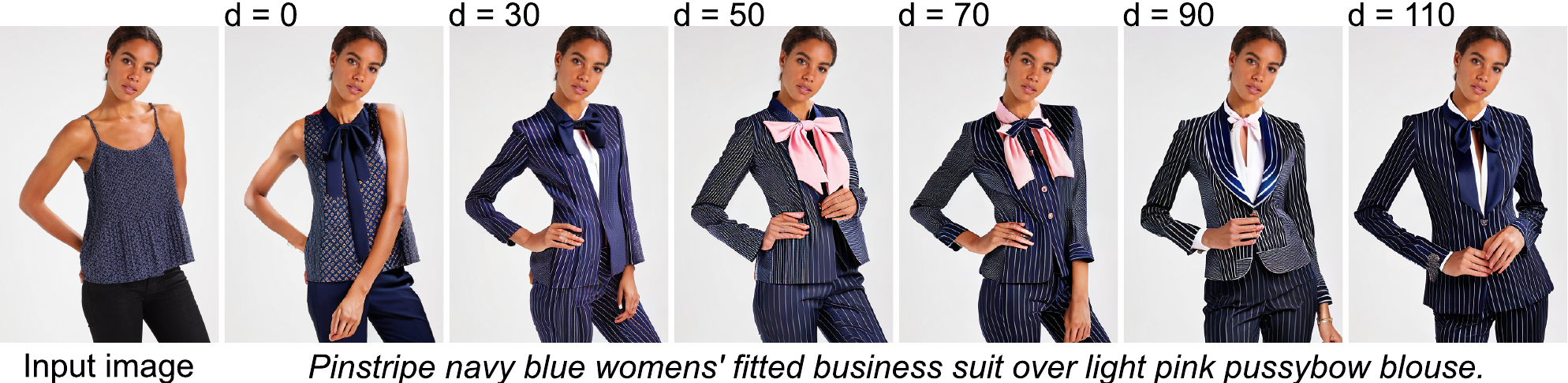}
    \caption{\textbf{Effect of mask dilation size on inpainting quality.} When too small, information on existing garments leaks into the model and forces them to incorporate it into the inpainted region, while setting it too large results in loss of detail.\vspace{-4mm}}
    \label{fig:ablation_study}
\end{figure}

\subsection{Ablation study}
We perform an ablation study with different sizes of the dilation kernel to evaluate its impact on the performance of DiCTI. We use four text prompts and generate five images for each of the test images from VITON-HD, yielding over 8000 generated images per kernel size. The scores for each of the kernel sizes are shown in Tab.~\ref{tab:ablation}. The table suggests that the lowest KID score is achieved for $d = 0$, meaning that the model just uses a raw binary mask, generated directly from the DensePose segmentation. The KID value gradually increases with an increasing dilation size. This may be attributed to the fact that the dataset is relatively small, so the distribution of the images with no mask dilation may be the most similar to that of the dataset images since the generated garments are forced to be more similar to the original ones that "leak" into the inpainting model without being masked out. Conversely, the CLIP score shows better performance for higher values of $d$. We speculate this is caused by more freedom with creating the garment, since the inpainting area is larger and the effect of garments in the original image diminishes. Interestingly, CLIP-IQA peaks at $d = 70$, which correlates with our initial empirical setting of the parameter. The differences are, however, relatively small, suggesting that dilation kernel size does not have a critical influence on the realism of the image as perceived by a deep neural model for higher values of $d$.  
A similar observation can be made for the Fashionpedia dataset. All of the metrics gradually increase. CLIP score and CLIP-IQA stabilize after 70, suggesting that the effect of increasing kernel size has past 70 diminishes.
The difference is furthermore visualized in Fig.~\ref{fig:ablation_study}. The result for $d = 0$ completely fails to generate the correct garment shape due to a restrictive inpainting mask. The result for $d = 30$ is better, but the sleeves appear slightly unnaturally tight. The results for $d > 50$ appear much better visually, but we notice that the pose change increases with the growing size of the kernel.

\begin{figure}[t]
\centering
\includegraphics[width=0.999\linewidth]{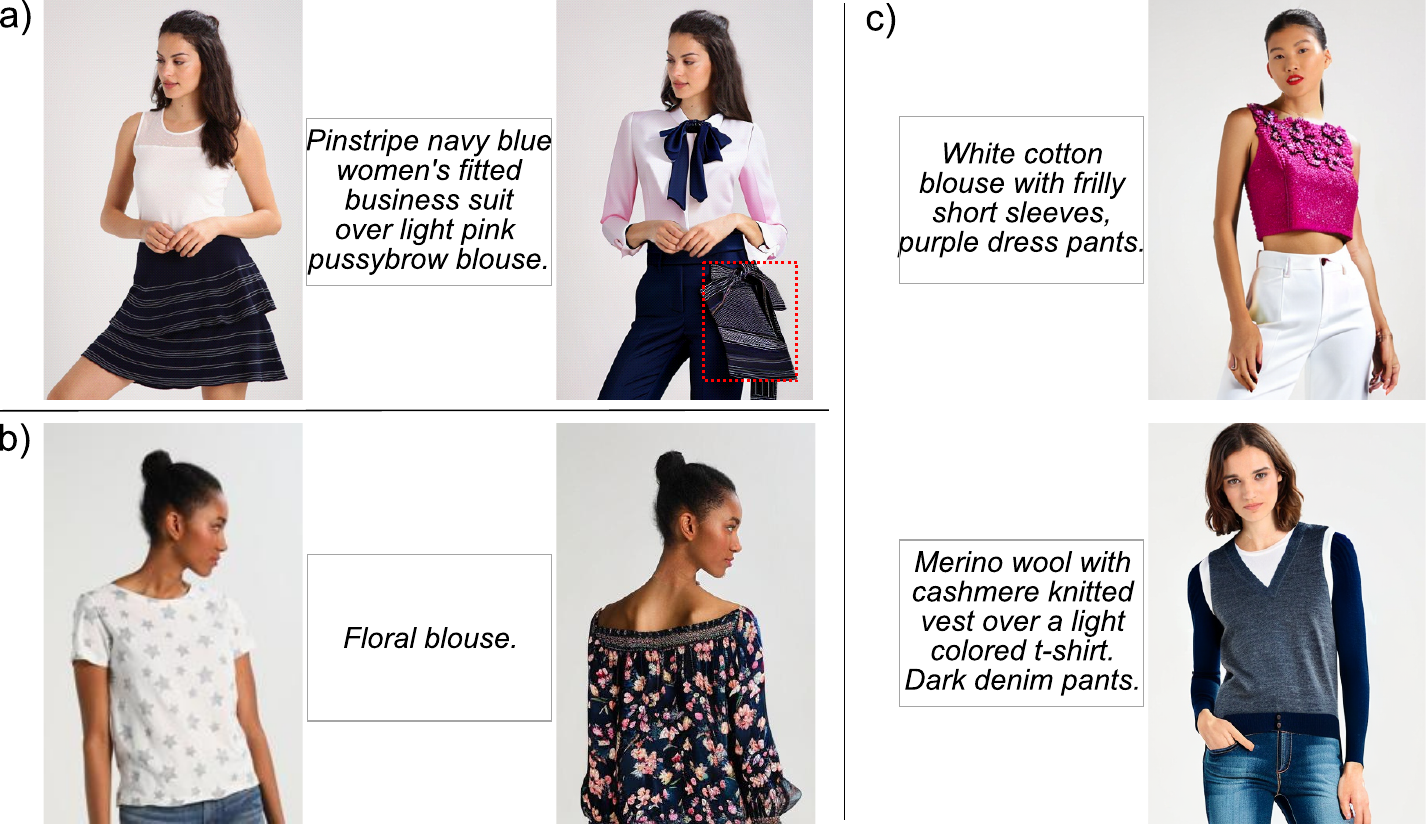}
\caption{\textbf{Limitations of DiCTI.} a) The inpainting mask fails to cover all existing clothes, b) the model fails to interpolate the pose of the covered area, and c) the model fails to understand the relations between parts of the text prompt.\vspace{-2mm}}
\label{fig:limitations}
\end{figure}

\subsection{Limitations}
As demonstrated in the previous sections, DiCTI achieves highly realistic results, faithful to the text descriptions. We do however notice a few scenarios in which the performance degrades a little.
Those can be roughly grouped into three groups, as shown in Fig.~\ref{fig:limitations}. 
Sometimes the automatically generated mask fails to cover the entirety of an existing garment. This is most noticeable in cases when the existing garment is very loose. Unlike many similar, masking-dependent methods, DiCTI generally still produces a realistic and fashionable garment image by incorporating the existing garment into the new one, sometimes adding a nice artistic touch to the image (Fig.~\ref{fig:limitations}a). 
Since the inpainting module gets little information about the body pose, save for the head and hands position, the pose tends to change a bit compared to the input image (Fig.~\ref{fig:limitations}b). Despite the changes, the resulting images depict people in realistic poses without degradations in image quality. 
Text-prompt faithfulness (like most similar methods) is bound to the underlying pre-trained language model. When a text prompt is very elaborate, containing many features of several different clothing items, the information sometimes gets mixed up among items (Fig.~\ref{fig:limitations}c). This can often be solved with a little prompt-engineering, which is becoming a very common technique in artistic applications of text-image models.


\addtolength{\textheight}{-5cm}   

\section{Conclusions}
In this work, we proposed a consumer-targeted pipeline for fashion item design, called DiCTI. 
DiCTI leverages the capabilities of the recently developed general-purpose diffusion-based generative models within a text-guided inpainting scheme. 
Despite its apparent simplicity, evaluation results show that DiCTI generates images of high visual quality and can cover a wide range of clothing shapes, materials, and colors; and that it clearly outperforms the previous state-of-the-art. As part of our future work, we plan to extend the approach to more general prompting input that in addition to text would also include other target characteristics.

{\small
\bibliographystyle{ieee}
\bibliography{egbib}
}

\end{document}